\documentclass[10pt,twocolumn,letterpaper]{article}

\usepackage{cvpr}              

\usepackage[dvipsnames]{xcolor}

\definecolor{cvprblue}{rgb}{0.21,0.49,0.74}
\usepackage[pagebackref,breaklinks,colorlinks,citecolor=cvprblue]{hyperref}
\usepackage{bbding}
\usepackage{color}
\usepackage{booktabs}
\usepackage{listings}
\def \pzo {\phantom{0}} 
\newcommand{\Epsilon}{\mathcal{E}} 
\usepackage{multicol}
\usepackage{multirow}
\usepackage{makecell}
\newcommand{\secref}[1]{$\S$\ref{#1}}
\def\paperID{10777} 
\def\confName{CVPR}
\def\confYear{2024}

\title{PortraitBooth: \\A Versatile Portrait Model for Fast Identity-preserved Personalization}

\author{Xu Peng$^{1}$, Junwei Zhu$^{2}$, Boyuan Jiang$^{2}$, Ying Tai$^{3}$, Donghao Luo$^{2}$, Jiangning Zhang$^{2}$, Wei Lin$^{1}$, Taisong Jin$^{1}$, \\Chengjie Wang$^{2}$, Rongrong Ji$^{1}$ \\
$^1$Xiamen University, $^2$Tencent, $^3$Nanjing University\\
\url{https://portraitbooth.github.io}
}

\begin{document}
\twocolumn[{%
\renewcommand\twocolumn[1][]{#1}%
\maketitle
\begin{center}
    \centering
    \captionsetup{type=figure}
    \includegraphics[width=1\textwidth]{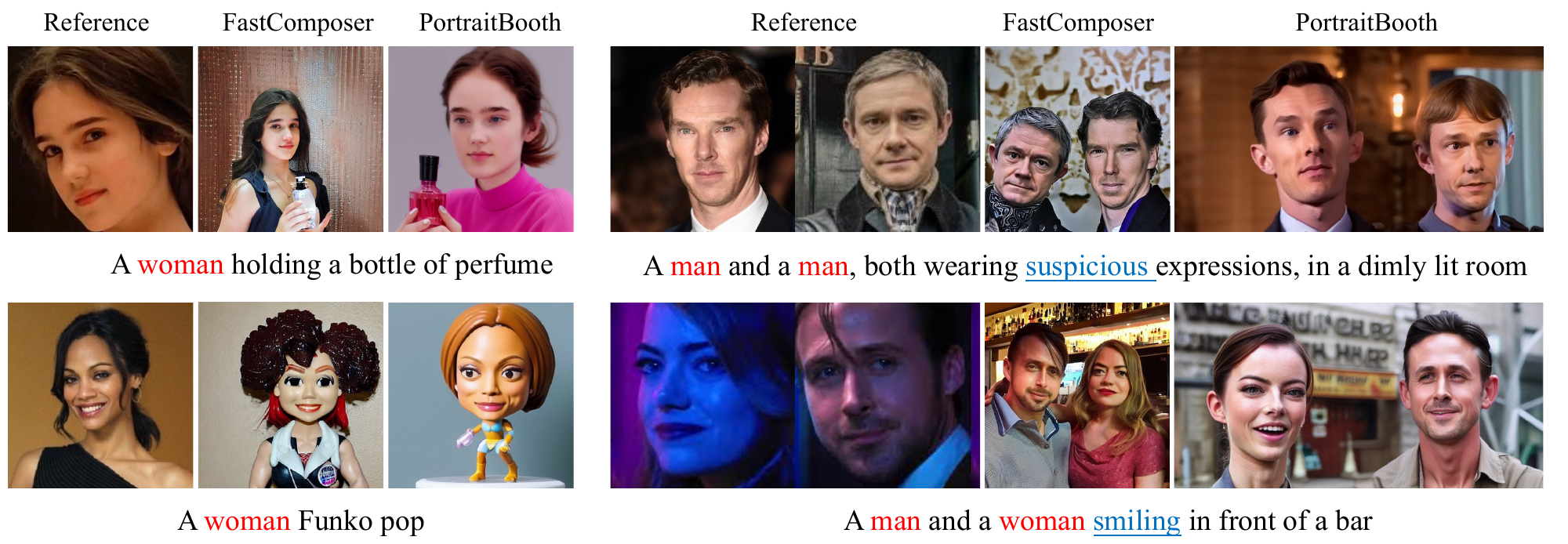}
    \vspace{-6mm}
    \caption{\textbf{Qualitative comparison of PortraitBooth and FastComposer} on action, style, expression editing, multi-subject generation, and identity preservation, all without any test-time tuning. }
    \label{figure1}
\end{center}%
}]
\begin{abstract}
Recent advancements in personalized image generation using diffusion models have been noteworthy.
However, existing methods suffer from inefficiencies due to the requirement for subject-specific fine-tuning.
This computationally intensive process hinders efficient deployment, limiting practical usability.
%
Moreover, these methods often grapple with identity distortion and limited expression diversity. 
In light of these challenges, we propose PortraitBooth, an innovative approach designed for high efficiency, robust identity preservation, and expression-editable text-to-image generation, without the need for fine-tuning. 
PortraitBooth leverages subject embeddings from a face recognition model for personalized image generation without fine-tuning.
It eliminates computational overhead and mitigates identity distortion. 
The introduced dynamic identity preservation strategy further ensures close resemblance to the original image identity.
%
%
Moreover, PortraitBooth incorporates emotion-aware cross-attention control for diverse facial expressions in generated images, supporting text-driven expression editing. 
Its scalability enables efficient and high-quality image creation, including multi-subject generation. 
Extensive results demonstrate superior performance over other state-of-the-art methods in both single and multiple image generation scenarios.
\end{abstract}

\section{Introduction}
\label{sec:intro}

Recent years have witnessed remarkable progress in text-to-image synthesis~\cite{avrahami2023spatext,kumari2023multi,wei2023elite,ramesh2021zero}, propelled by the emergence of diffusion models~\cite{ho2020denoising,ho2022classifier,rombach2022high,chefer2023attend,xue2023raphael}. Pre-trained text-to-image generation models have opened up new avenues for creative content creation, with personalized generation gaining popularity for its diverse applications.

\begin{figure}[t!]
	\centering
	\includegraphics[width=0.46\textwidth]{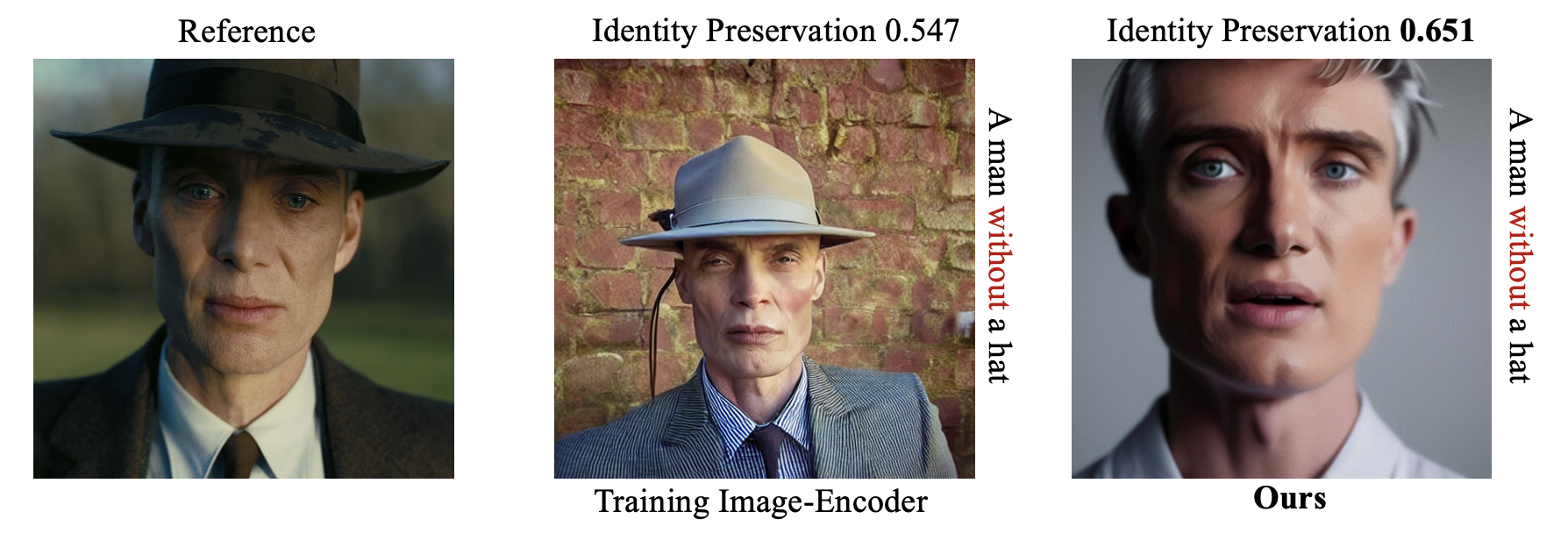}
	\vspace{-3mm}
	\caption{Comparison of identity information obtained based on the trained image encoder and pre-trained face recognition model.}
	\label{clip_vs_us}
\end{figure}

%
%
%
Personalized generation methods based on diffusion models fall into two main categories: \textbf{\textit{1)}} test-time fine-tuning and \textbf{\textit{2)}} test-time non-fine-tuning. 
Some approaches~\cite{ruiz2023dreambooth,gal2022image,nitzan2022mystyle,hao2023vico,ruiz2023hyperdreambooth} endorse test-time fine-tuning using reference images (typically $3$-$5$) to generate personalized results. 
However, these methods require specialized network training~\cite{ruiz2023dreambooth,smith2023continual} , making them \textit{inefficient for practical applications}.
An alternative to test-time fine-tuning is retraining the base text-to-image model with specially designed strategies, \eg training a distinct image encoder on a massive dataset to capture reference image identity information. 
However, these approaches~\cite{ma2023subject,xiao2023fastcomposer,valevski2023face0} face challenges, either \textit{dealing with identity distortion} or \textit{generating images lacking editability}, as depicted in Fig.~\ref{clip_vs_us}. 
This is mainly due to the coarse-grained nature of the identity information obtained from the trained image encoder. 
The better the image encoder is trained, the tighter the identity information with reference image is coupled, severely compromising editability. 
Additionally, these methods often demand significant GPU resources and high storage, making them impractical for most research institutions.
%
\cref{table:table1} offers a comprehensive comparison of existing personalized image generation methods across four key aspects.

In this paper, we introduce PortraitBooth, a novel text-to-portrait personalization framework that achieves high efficiency, robust identity preservation, and diverse expression editing. 
We then describe our main characteristics in detail:

\noindent\textbf{High Efficiency.}
PortraitBooth stands out as a highly efficient one-stage generation framework, delivering the following advantages:
\textbf{\textit{1)}} Only a \textit{single} image is required during the inference stage, unlike other schemes such as Dreambooth that need multiple images. 
\textbf{\textit{2)}} \textit{No finetuning or optimization} is conducted during inference, which saves time and avoids delays. 
\textbf{\textit{3)}} \textit{Lower training resource requirement} is needed than Face0 and Subject-Diffusion that demand a lot of high-performance GPU resources.

\begin{table}[t!]
	\centering
	\resizebox{0.95\linewidth}{!}{
		\begin{tabular}{c|cccc}
			\toprule
			\multirow{2}{*}{\makecell[c]{\bf{Methods}}} & \multirow{2}{*}{\makecell[c]{\bf{Single Image}}} & \multirow{2}{*}{\makecell[c]{\bf{Test-time} \\ \bf{None-fine-tuning}}} & \multirow{2}{*}{\makecell[c]{\bf{Robust ID}\\ \bf{Preservation}}} & \multirow{2}{*}{\makecell[c]{\bf{Expression}\\ \bf{Editing}}} \\
			& & & & \\
			\midrule
			Textual Inversion~\cite{gal2022image} & \color{red}\XSolidBrush & \color{red}\XSolidBrush & \color{red}\XSolidBrush & \color{green}\Checkmark \\
			Dreambooth~\cite{ruiz2023dreambooth} & \color{red}\XSolidBrush & \color{red}\XSolidBrush & \color{red}\XSolidBrush & \color{green}\Checkmark \\
			Custom Diffusion~\cite{kumari2023multi} &\color{red}\XSolidBrush & \color{red}\XSolidBrush & \color{red}\XSolidBrush & \color{green}\Checkmark \\
			Break-A-Scene~\cite{avrahami2023break} &\color{green}\Checkmark & \color{red}\XSolidBrush & \color{green}\Checkmark & \color{red}\XSolidBrush \\
			HyperDreamBooth~\cite{ruiz2023hyperdreambooth} &\color{green}\Checkmark & \color{red}\XSolidBrush & \color{green}\Checkmark & \color{green}\Checkmark\\
			FastComposer~\cite{xiao2023fastcomposer} &\color{green}\Checkmark & \color{green}\Checkmark & \color{red}\XSolidBrush & \color{red}\XSolidBrush\\
			Face0~\cite{valevski2023face0} &\color{green}\Checkmark & \color{green}\Checkmark & \color{green}\Checkmark & \color{red}\XSolidBrush\\
			Subject-Diffusion~\cite{ma2023subject} &\color{green}\Checkmark & \color{green}\Checkmark & \color{green}\Checkmark & \color{red}\XSolidBrush\\
			\midrule
			\bf{PortraitBooth (Ours)}  &\color{green}\Checkmark & \color{green}\Checkmark & \color{green}\Checkmark & \color{green}\Checkmark\\
			\bottomrule
		\end{tabular}
	}
	\vspace{-2mm}
	\caption{\textbf{Comparisons of current personalization approaches}. 
	}
	\vspace{-0.5cm}
	\label{table:table1}
\end{table}

\noindent\textbf{Robust Identity Preservation.}
\textbf{\textit{1)}} 
PortraitBooth employs a pre-trained face recognition model ($41.5$M parameters) to 
extract a face embedding from a given reference image. 
This embedding is then projected into the context space of Stable Diffusion using a simple multilayer perceptron, enabling high-fidelity image generation based on the proposed Subject Text Embedding Augmentation (STEA).
\textbf{\textit{2)}} PortraitBooth Dynamically maintains Identity Preservation (DIP) by incorporating an identity loss during training to facilitate the model to ensure identity preservation.

\noindent\textbf{Diverse Expression Editing.}
While the discriminative features extracted from a robust face recognition model effectively disentangle identity and attributes, expression editing remains a challenge for existing one-shot methods~\cite{ma2023subject}.
To address this, we introduce Emotion-aware Cross-Attention Control (ECAC) via a truncation mechanism. This allows a single area to respond to multiple tokens simultaneously, thereby enabling versatile expression editing (see Fig.~\ref{figure1}). 

In summary, our contributions are threefold:
\begin{itemize}
\item We propose a novel one-shot text-to-portrait generation framework, termed PortraitBooth, which is the first solution to achieve high efficiency, robust identity preservation, and low training cost, simultaneously.



\item 
To address identity distortion, we introduce the STEA and DIP modules for robust identity preservation. Additionally, we propose the ECAC module, achieving diverse expression editing.
\item 
Our method scales effortlessly for single-subject and multi-subject generation, integrating smoothly with multi-object generation methods. 
Furthermore, our PortraitBooth excels in achieving remarkable fidelity and editability, surpassing other state-of-the-art methods.

\end{itemize}


\section{Related Work}
\label{sec:relate work}

\paragraph{Image Editing with Diffusion Models.}

Image editing \cite{tumanyan2023plug,ge2023expressive} is a fundamental task in computer vision, involving modifications to an input image with auxiliary inputs like audio~\cite{zhang2021real,zhang2023sadtalker}, text~\cite{xu2023high}, masks~\cite{ge2023expressive}, or reference images~\cite{zhang2020freenet,xu2022styleswap,xu2022designing,xu2022region}. 
Despite the capabilities of large-scale diffusion models such as Imagen \cite{saharia2022photorealistic}, DALL·E2 \cite{ramesh2022hierarchical}, and Stable Diffusion \cite{rombach2022high} in text-to-image synthesis, they lack precise control over image generation solely through text guidance. 
Even a small change in the original prompt can yield significantly different outcomes.
Recent research has focused on adapting text-guided diffusion models~\cite{avrahami2022blended,couairon2022diffedit,kim2022diffusionclip,kawar2023imagic,hertz2022prompt,hu2021lora} for real image editing, leveraging their rich and diverse semantic knowledge. 
One such approach is Prompt-to-Prompt~\cite{hertz2022prompt}, which injects internal cross-attention maps when modifying only the text prompt, preserving the spatial layout and geometry necessary for regenerating an image while modifying it through prompt editing. 
Existing methods for portrait expression editing based on diffusion models not only focus on designing optimization-free methods~\cite{meng2021sdedit,choi2021ilvr,avrahami2023blended,preechakul2022diffusion}, but also explore face swapping as an alternative approach. For example, 
DiffusionRig~\cite{ding2023diffusionrig} learns generic facial personalized priors to control face synthesis.

\paragraph{Personalized Visual Content Generation.}

Personalized visual content generation aims to create images tailored to individual preferences or characteristics, including new subjects described by one or more images~\cite{gal2023encoder}. Textual Inversion (TI)~\cite{gal2022image} and DreamBooth (DB)~\cite{ruiz2023dreambooth} are two pioneering works in personalization. They generate different contexts for a single visual concept using multiple images. TI introduces a learnable text token and optimizes it for concept reconstruction using standard diffusion loss, while keeping model weights frozen. DB reuses a rare token and fine-tunes model weights for reconstruction. HyperDreamBooth~\cite{ruiz2023hyperdreambooth} offers a lightweight, subject-driven personalization for text-to-image diffusion models compared to DB. 
Custom Diffusion~\cite{kumari2023multi} fine-tunes subset layers of the cross-attention in the UNet. However, these tuning-based methods require time-consuming fine-tuning or multiple images, which is inelegant. In contrast, PortraitBooth amortizes costly subject tuning during training, enabling fast personalization with a single image.

Concurrent tuning-free methods include~\cite{xiao2023fastcomposer,valevski2023face0,ma2023subject}, those use an image encoder for accessibility, but Fastcomposer may distort identity due to lack of fine-grained training. Face0~\cite{valevski2023face0} and Subject-Diffusion~\cite{ma2023subject} achieve relatively high identity preservation in personalized generation through massive datasets and expensive hardware resources. However, they require resource-intensive backpropagation. Conversely, PortraitBooth generates personalized portraits with comparable identity preservation in an inference-only manner, requiring fewer hardware resources that most research institutions can afford.
\begin{figure*}
    \centering
    \includegraphics[width=0.8\textwidth]{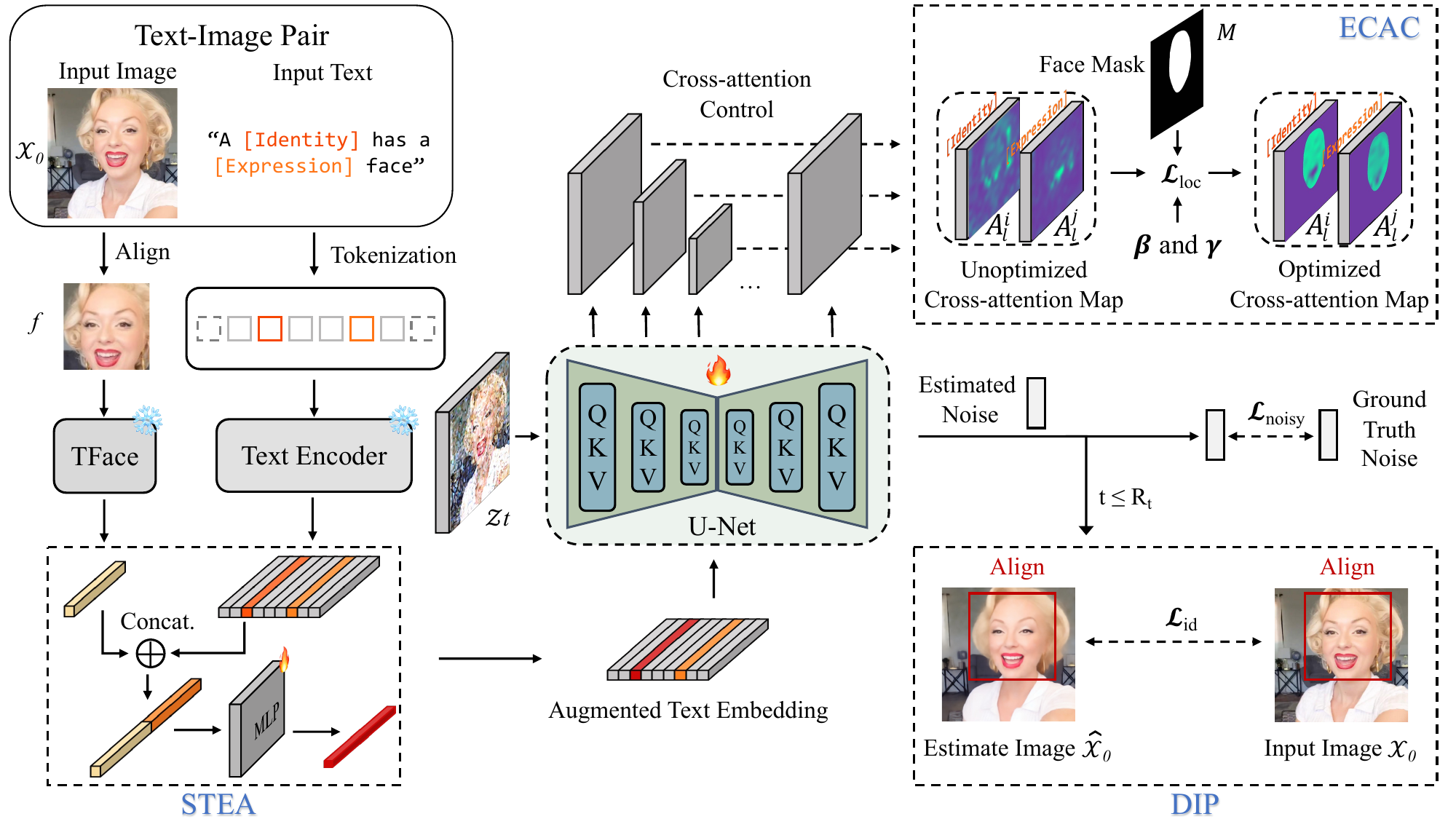}
    \vspace{-2.5mm}
    \caption{\textbf{Overview framework of PortraitBooth}. 
    %
    PortraitBooth extracts the face $f$ from the input image $x_0$, and augments the subject's features using TFace for improved identity representation. 
    The diffusion model is trained to generate images with enhanced conditioning, incorporating emotion-aware cross-attention for expression editing and dynamic identity preservation to maintain identity. 
    During the testing phase, we only need to input a single image and the corresponding prompt to achieve rapid, robust identity preservation and diverse expression editing capabilities. 
    %
    $A^i_l$, $A^j_l$  represents the cross-attention map corresponding to the $i$-th and $j$-th token at the $l$-th cross-attention layer, respectively. 
    $\beta$ and $\gamma$ represent the maximum values of the cross-attention map for the identity token and expression token respectively, while $R_t$ indicates the timing for identity preservation.
    }
    \label{mainframework}
\vspace{-0.5cm}
\end{figure*}
\section{Preliminaries}
\subsection{Stable Diffusion}


Stable Diffusion (SD) consists of three components: a Variational AutoEncoder (VAE), a conditional U-Net~\cite{ronneberger2015u}, and a text encoder~\cite{radford2021learning}. Specifically, for an input image $x_0$, The VAE encoder $\Epsilon$ compresses the $x_0$ to a smaller latent representation $z$. The diffusion process is then performed on the latent space, where a conditional U-Net denoiser $\epsilon_{\theta}$, denoises the noisy latent representation by predicting the noise $\theta$ with current timestep $t$,  $t$-th noisy latent $z_{t}$. 
This denoising process can be conditioned on textual conditional $C$ through the cross-attention mechanism, 
Throughout the training process, the network is optimized to minimize the loss function defined as:
\begin{equation}
\begin{split}
\mathcal{L}_{noise} &= \mathbb{E}_{z \sim \Epsilon(x), C, \epsilon \sim \mathcal{N}(0,1), t} \left[||\epsilon - \epsilon_{\theta}(z_t,t,C)||^2_{2} \right], \\\\
 &\;\quad\quad\quad z_t \sim \mathcal{N}(\sqrt{\alpha_t}z_{t-1}, 1-\alpha_t),
\end{split}
\end{equation}
where $\alpha_{t}$ is a predefined sequence of coefficients controlling the variance schedule. 
The closed form of the distribution $p(z_t|z_0)$ can be easily derived as:
\begin{equation}
\begin{split}
z_t &= \sqrt{\Bar{\alpha}_{t}}z_0 + (1-\Bar{\alpha}_{t})\epsilon,\\ 
\bar{\alpha}_{t} &= \prod_{s=1}^{t}\alpha_s, \epsilon \sim \mathcal{N}(0, 1).
\end{split}
\label{one-step-reference}
\end{equation}

\subsection{Cross-Attention Mechanism}
In the SD model, the U-Net employs a cross-attention mechanism to denoise the noisy latent image conditioned on text prompts. 
The cross-attention layer accepts the spatial noisy latent image $z_t$ 
and the text embeddings $y$ as inputs. 
The embeddings of the visual and textual features are fused to produce spatial attention maps for each textual token. The cross-attention maps are computed with:

\begin{equation}
A = softmax\left(\frac{QK^T}{\sqrt{d}}\right). \label{cross-attention-score}
\end{equation}
The query matrix, denoted as $Q$ = $z_tW^{(i)}_{Q}$, is the projection of the noisy latent image $z_t$. The key matrix, represented as $K$ = $yW^{(i)}_{K}$, is the projected textual features. Here, $W^{(i)}_{Q}$ and $W^{(i)}_{K}$  represent the weight matrices of the two linear layers in each cross-attention block $i$ of the U-Net, and $d$ is the output dimension of $K$ and $Q$ features. 



\section{Methodology}
\subsection{Subject Text Embedding Augmention}
From a generative standpoint, our objective is to create a portrait that accurately represents the identity of the source face. To achieve this, we utilize a pre-trained face recognition model called TFace~\cite{huang2020curricularface} to extract the identity features. In order to better preserve the identity, we incorporate face features as an important input condition and integrate them into the text to enhance its ability to capture the nuances of identity. To elaborate, we first encode the text prompt $P = \{w_{1},w_{2},...w_{n}\}$ and reference face $f$ into embeddings using the pre-trained text encoder and TFace, denoted as $\psi$ and $\varphi$ respectively. However, as the features generated by the recognition model are primarily designed for recognition purposes and may not be optimal for generation, we choose to extract only the \textit{shallow features} of the recognition model. Subsequently,
we concatenate the embedding of the identity token with the facial feature, and then feed the resulting augmented embeddings into the $MLP$. This process yields the final conditioning embeddings $C = \{c_1,c_2...c_n\}$, which are defined as :
\begin{equation}
    \resizebox{0.9\hsize}{!}{$
        c_i = \begin{cases}
        \psi(w_i) &  w_i \not\in \{identity\quad token\} \\
        MLP([\psi(w_i) || \varphi(f)]) & w_i \in \{identity\quad token\}.
        \end{cases}
        $}
\end{equation}
This approach allows us to generate portraits that not only capture the textual description but also incorporate the identity features extracted from the reference face, resulting in a more accurate representation of the desired identity. Fig.~\ref{mainframework} illustrates the STEA module, which provides a concrete example of our augmentation approach. 

\subsection{Dynamic Identity Preservation}
The current SD model achieves image fidelity by relying on accurate prompts, which however poses a significant challenge.
When incorporating new image conditions, ensuring the fidelity of the unique reference image becomes necessary. 
Therefore, it is crucial to incorporate identity loss into the training framework of diffusion models to ensure identity preservation. Let $x_0$ be the input image, $z$ be its latent space representation, $T$ ($T <$ 1000) represents the total number of noise injection steps. For a small value of $R_t$ ($R_t < T$), we can get estimated $\hat{z}_0$ directly from $z_t$ and the predicted noise $\epsilon_{\theta}(z_t,t,C)$. From Eqn. \ref{one-step-reference}, the one-step reverse formula is defined as :
\begin{equation}
    \hat{z}_0 = \dfrac{z_t-\sqrt{1-\bar{\alpha}_t}\epsilon_{\theta}}{\sqrt{\bar{\alpha}_t}}, t \leq R_t,
\end{equation}
After reverse, the estimated $\hat{z}_0$ is decoded from the latent space using the VAE decoder $\mathcal{D}$ to obtain the estimated input image $\hat{x}_0 = \mathcal{D}(\hat{z}_0)$. Then, based on the facial region of the original image, the estimated facial region image $\hat{x}^{f}_0$ is extracted from the reconstructed image. Finally, the identity loss between the estimated facial image and the reference facial image is defined as:
\begin{equation}
\mathcal{L}_{id} = \begin{cases} 1 - CosSim\left(\varphi(f),\varphi(\hat{x}_0^{f})\right) & t \leq R_t\\
0 & t > R_t.
\end{cases}
\end{equation}
The identity loss is designed to handle noisy images and improve the model's ability to preserve the identity. The DIP module, as illustrated in Fig.~\ref{mainframework}.

\begin{figure*}
	\centering
	\includegraphics[width=0.82\textwidth]{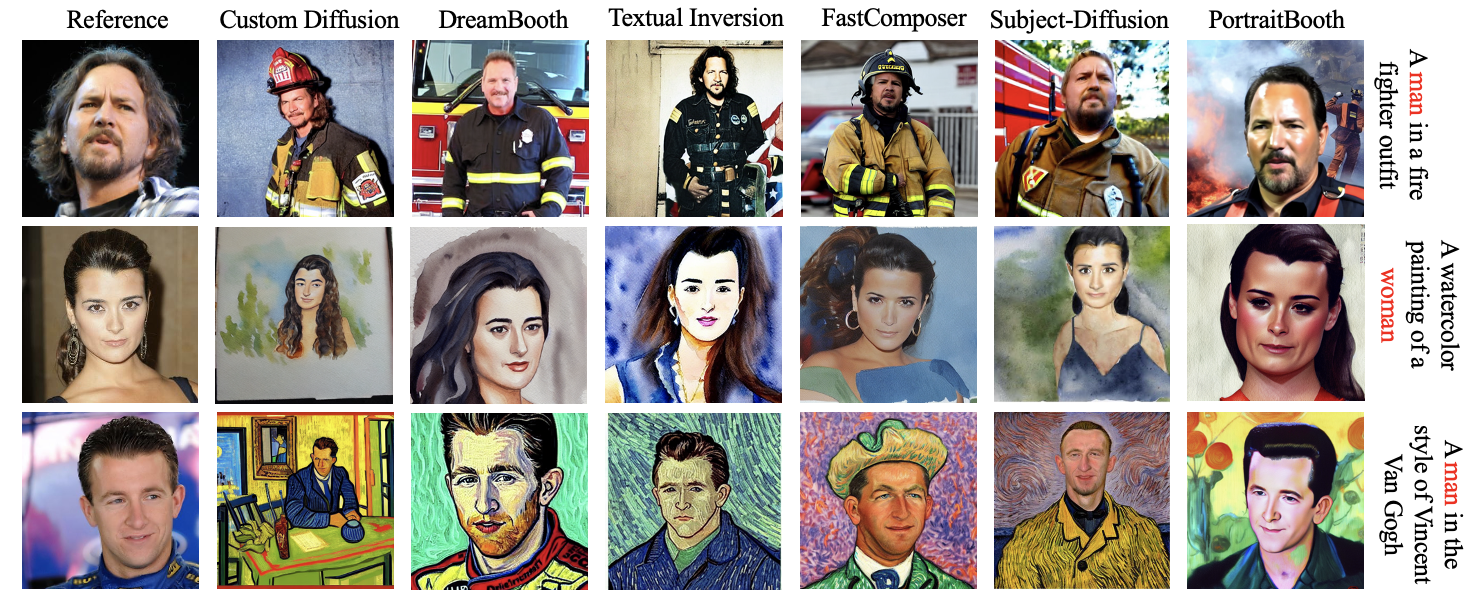}
	\vspace{-4mm}
	\caption{\textbf{Comparison of different methods on single subject image generation} in the testing dataset.}
	\label{personcmp}
\end{figure*}

\begin{figure}
	\centering
	\includegraphics[width=0.51\textwidth]{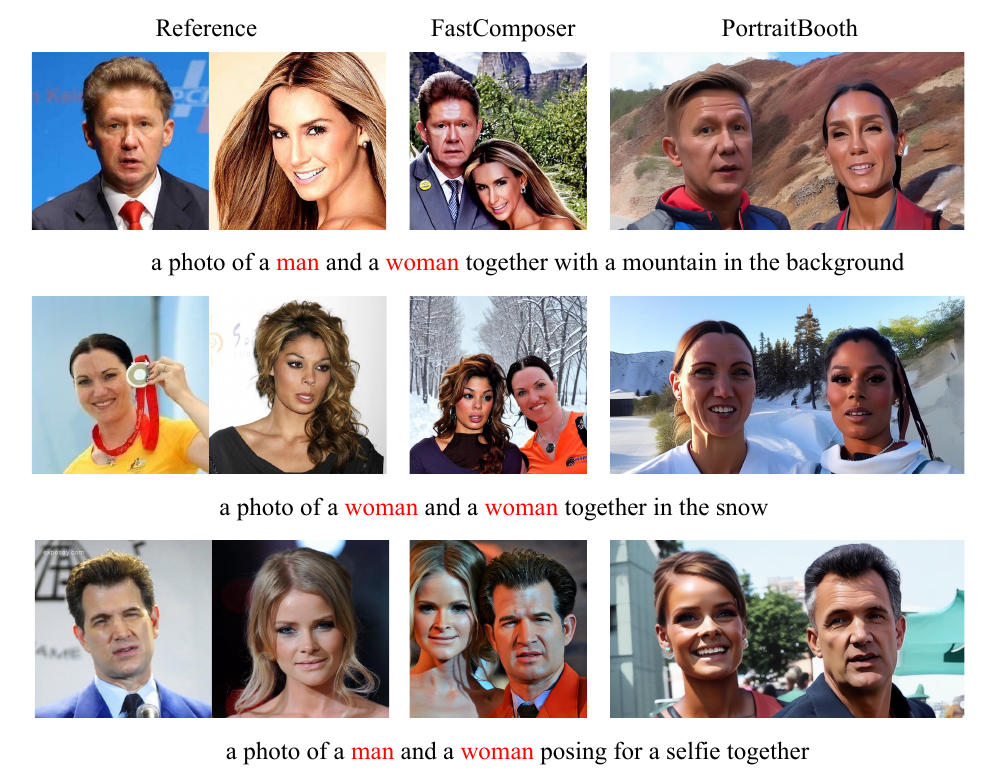}
	\vspace{-7mm}
	\caption{\textbf{Comparison of different methods on multi-subject image generation} in the testing dataset.}
	\label{multi-subject}
\end{figure}

\subsection{Emotion-aware Cross-attention Control}
For previous one-shot personalized generation works~\cite{ma2023subject,xiao2023fastcomposer,valevski2023face0} , a common issue is that the generated images always have the same expression as the reference image, regardless of the prompt given. Although we have largely decoupled identity and attributes by utilizing pre-trained facial recognition models to extract discriminative features for subject feature enhancement, the complexity and diversity of facial expressions still pose a challenge in maintaining identity during portrait generation. This issue primarily arises because the cross-attention map is spread across the entire image during image generation. To address this issue and ensure that the cross-attention map corresponding to specific tokens only attends to the image region occupied by the corresponding concept, we propose an emotion-aware cross-attention control mechanism.

Specifically, unlike previous methods \cite{xiao2023fastcomposer,ma2023subject} that used attention masks to control 
subject token’s attention map solely on the one subject region, we allow attention control of different tokens within the same region by truncating cross-attention mechanism. 
For instance, when dealing with tokens for facial expressions and identity, we employ a face mask to ensure that the attention maps corresponding to these two tokens are both focused on the face region. However, we observe that when two different tokens' attention maps are both constrained to the same region, one token may learn well while the other may not. 
To tackle this problem, we propose a complete local control constraint with truncating cross-attention mechanism:
\begin{equation}
\setlength{\abovedisplayskip}{3pt}
\setlength{\belowdisplayskip}{3pt}
\resizebox{0.9\hsize}{!}{$
\begin{split}
   \mathcal{L}_{loc} &= \dfrac{1}{N}\sum_{l=1}^N\lambda(mean(A^i_l(1-M))+mean(relu(\beta-A^i_l)M)) \\
     &+ \dfrac{1}{N}\sum_{l=1}^N\mu(mean(A^j_l(1-M))+mean(relu(\gamma-A^j_l)M)),
     \end{split}
     $}
\end{equation}
where $M$ is the face mask normalized to [0,1]. $mean$ is the pixel-level averaging. $A^i_l, A^j_l \in [0,1]$ represents the cross-attention map corresponding to the identity and expression token at the $l$-th cross-attention layer. $\beta$ and $\gamma$ are used to constrain the maximum response intensity of the cross-attention map in the facial area corresponding to the identity token and expression token, respectively.
We optimize $\mathcal{L}_{loc}$ to ensure that objects' attention map exhibit their respective response in the desired area, which is achieved by maximizing the response of each token's attention map to the face region and minimizing its response to the background, along with the use of a truncated response mechanism in the attention map. $\lambda$ and $\mu$ are localization loss ratios, which are $0.001$ and $0.01$. Fig.~\ref{mainframework} illustrates the ECAC module.


\begin{table*}[t]
	\begin{tabular}{ccccccc}
		\toprule
		\bf{Method} & \bf{Type} & \bf{Reference Image} $\downarrow$ & \bf{Id Pres.} $\uparrow$ & \bf{CLIP-TI} $\uparrow$ & \bf{Test Time} $\downarrow$ & \bf{Training Cost} \\
		\midrule
		Stable Diffusion \cite{rombach2022high} & Zero Shot & 0 & 0.039 & 0.268 & $\approx$2s\pzo\pzo\pzo & -   \\
		Face0 \cite{valevski2023face0} & One Shot & 1 & - & - & $\approx$2s\pzo\pzo\pzo & 64 TPU   \\
		\midrule
		Textual-Inversion \cite{gal2022image} & Finetune & 5 & 0.293 & 0.219 & $\approx$2500s & 1 A100   \\
		DreamBooth \cite{ruiz2023dreambooth} & Finetune & 5 & 0.273 & 0.239 & $\approx$1084s & 1 A100   \\
		Custom Diffusion \cite{kumari2023multi} & Finetune & 5 & 0.434 & 0.233 & $\approx$789s\pzo & 1 A100   \\
		FastComposer \cite{xiao2023fastcomposer} & One Shot & \bf{1} & 0.514 & 0.243 & $\approx$\bf{2s}\pzo\pzo\pzo & 8 A6000   \\
		Subject-Diffusion \cite{ma2023subject} & One Shot & \bf{1} & 0.605 & 0.228 & $\approx$\bf{2s}\pzo\pzo\pzo & 24 A100   \\
		\bf{PortraitBooth (ours)} & One Shot & \bf{1} & \bf{0.657} & \bf{0.245} & $\approx$\bf{2s}\pzo\pzo\pzo & 3 A100   \\
		\bottomrule
	\end{tabular}
	\vspace{-2mm}
	\caption{\textbf{Comparison between our method and baseline approaches on single-subject image generation.} Our approach achieves highly satisfactory results with the utilization of relatively limited resources under the one-shot setting.}
	\label{exps}
\end{table*}

\begin{table*}[t]
	\begin{tabular}{ccccccc}
		\toprule
		\bf{Method} & \bf{Type} & \bf{Reference Image} $\downarrow$ & \bf{Id Pres.} $\uparrow$ & \bf{CLIP-TI} $\uparrow$ & \bf{Test Time} $\downarrow$ & \bf{Training Cost} \\
		\midrule
		Stable Diffusion \cite{rombach2022high} & Zero Shot & 0 & 0.019 & 0.284 & $\approx$2s\pzo\pzo\pzo & -   \\
		\midrule
		Textual-Inversion \cite{gal2022image} & Finetune & 5 & 0.135 & 0.211 & $\approx$4998s & 1 A100   \\
		Custom Diffusion \cite{kumari2023multi}& Finetune & 5 & 0.054 & \bf{0.258} & $\approx$789s\pzo & 1 A100   \\
		FastComposer \cite{xiao2023fastcomposer} & One Shot & \bf{2} & 0.431 & 0.243 & $\approx$\bf{2s}\pzo\pzo\pzo & 8 A6000   \\
		\bf{PortraitBooth (ours)} & One Shot & \bf{2} & \bf{0.647} & 0.239 & $\approx${18s}\pzo\pzo & 3 A100   \\
		\bottomrule
	\end{tabular}
	\vspace{-2mm}
	\caption{\textbf{The comparison between our method and the baseline approaches that support multiple-subject image generation.}  StableDiffusion was used as the text-only baseline without any subject conditioning. 
	}
	\vspace{-0.4cm}
	\label{exps_multi}
\end{table*}

\subsection{Objective Function}
First, we use TFace $\varphi$ to extract the face embedding and concatenate it with specific identity token embedding, which is extracted from the text encoder $\psi$. These are then fed into an $MLP$ for feature enhancement, forming U-Net aware conditional information $C$. Next, we feed the noisy latent space feature map $z_t$ into a U-Net with conditional guidance to predict noise, while implementing a truncation mechanism for local attention control for specific tokens. To better preserve identity, we employ dynamic identity preservation method to calculate the loss between the estimated face image $\hat{x}^{f}_0$ and reference face $f$.
The final training objective of PortraitBooth is:
\begin{equation}
\setlength{\abovedisplayskip}{3pt}
\setlength{\belowdisplayskip}{3pt}
    \mathcal{L}_{total} = \mathcal{L}_{loc} + \mathcal{L}_{noise} + \mathcal{L}_{id}.
\end{equation}

\section{Experiments}
\subsection{Experimental Setups}
\noindent \textbf{Dataset Description.}
We constructed a single subject image-text paired dataset based on the CelebV-T dataset~\cite{yu2023celebv}, which consists of $70,000$ videos. To utilize the additional textual descriptions provided by CelebV-T, we randomly extracted the first or last frames of each video. Additionally, we used the Recognize Anything model~\cite{zhang2023recognize} to generate captions describing the main subject for all images. To enhance the robustness of our models, we randomly selected a frame from the middle section of each video and used the facial region as our reference face image. We employ the pre-train face parsing model~\cite{lee2020maskgan} to generate subject face segmentation masks for each image. \\
\noindent \textbf{Training Details.}
We start training from the Stable Diffusion v$1-5$~\cite{rombach2022high} model. To encode the identity inputs, we use TFace model. During training, we only train the U-Net, the MLP module. We train our models for $150$k steps on $6$ NVIDIA V$100$ GPUs (For the sake of easy and intuitive comparison later, we roughly convert $6$ NVIDIA V$100$ GPUs into $3$ NVIDIA A$100$ GPUs.), with a constant learning rate of $1e-5$ and a batch size of $2$. We train the model solely on text conditioning with $10$\% of the samples to maintain the model’s capability for text-only generation. 
To facilitate classifier-free guidance sampling~\cite{ho2022classifier}, we train the model without any conditions on 10\% of the instances. 
During training, we apply the loss only in the subject's face region to half of the training samples to enhance generation quality in the subject area. 
There are $11$ emotion words involved in truncating cross-attention control, such as happy, angry, sad, \textit{etc}. We select a value of $250$ for $R_t$ to obtain 
$\hat{z_{0}}$ through reverse. The selected identity label is from the categories \{``man",``woman"\}. During inference,  We use Euler~\cite{karras2022elucidating} sampling with $50$ steps and a classifier-free guidance scale of $5$ across all methods.\\
\noindent \textbf{Evaluation Metric.}
We evaluate the quality of image generation based on identity preservation (Id Pres.) and CLIP text-image consistency (CLIP-TI). Identity preservation is determined by detecting faces in the reference and generated images using MTCNN~\cite{zhang2016joint}, and then calculating pairwise identity similarity using FaceNet~\cite{schroff2015facenet}. 
For multi-subject evaluation, we identify all faces within the generated images and use a greedy matching procedure between the generated faces and reference subjects. For the evaluation of expression editing, we calculate the text-image consistency between the emotion words in each prompt and the corresponding generated images as our expression coefficient metric. For efficiency evaluation, we consider the total time for customization, including fine-tuning (for tuning-based methods) and inference.
We also take into consideration the total number of GPUs required throughout the entire procedure. All baselines, by default, are run with the standard set of hyperparameters as mentioned in their paper.
\subsection{Personalized Image Generation}\label{Personalized}
To evaluate our model's effectiveness in this area, we use the single-entity evaluation method employed in FastComposer~\cite{xiao2023fastcomposer} and compare our model's performance to that of other existing methods including 
DreamBooth~\cite{ruiz2023dreambooth}, Textual-Inversion~\cite{gal2022image}, Custom Diffusion~\cite{kumari2023multi}, and Subject-Diffusion~\cite{ma2023subject}. Methods \cite{ruiz2023dreambooth,gal2022image,kumari2023multi} were used the implementation from diffusers library \cite{von-platen-etal-2022-diffusers}. Considering that Face0~\cite{valevski2023face0} does not provide open-source code, we can only list the hardware resources mentioned in their paper as a point of comparison. Stable Diffusion~\cite{rombach2022high} was used as the text-only baseline. The entire test set comprises 15 subjects, and 30 texts. The evaluation benchmark developed a broad range of text prompts encapsulating a wide spectrum of scenarios, such as recontextualization, stylization, accessorization, and diverse actions. Five images were utilized per subject to fine-tune the optimization-based methods. For the one-shot method, a single randomly selected image was employed for each subject. As shown in Tab.~\ref{exps}, 
PortraitBooth significantly outperforms all baseline approaches in identity preservation. 
Fig.~\ref{personcmp} shows the qualitative results of single-subject personalization comparisons, employing different approaches across an array of prompts. 

\begin{figure}[t]
	\centering
	\includegraphics[width=0.47\textwidth]{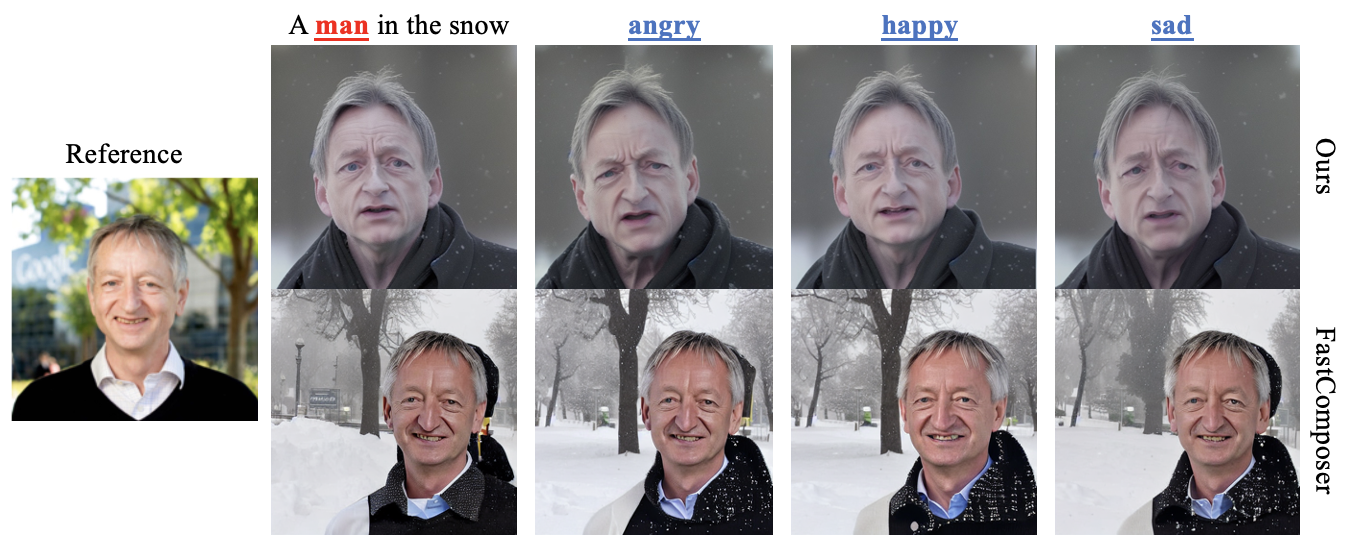}
	\vspace{-3mm}
	\caption{\textbf{Comparison chart of expression editing between our method and FastComposer,} focusing on the three most distinct expression terms.} 
	
	\label{emotioncmp}
\end{figure}

\begin{figure}[t!]
	\centering
	\includegraphics[width=0.45\textwidth]{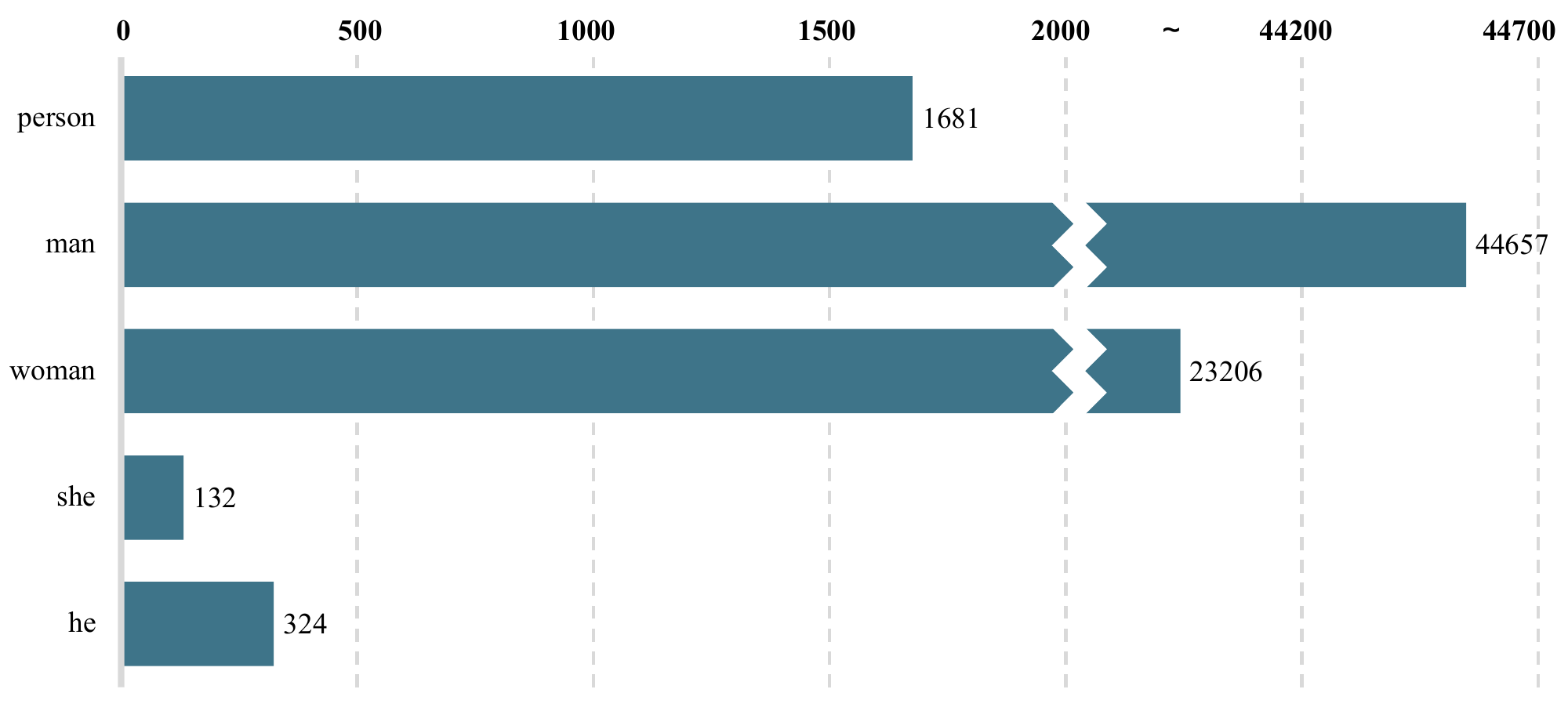}
	\vspace{-3mm}
	\caption{\textbf{The number of main subject words} occurrences in the generated $70,000$ captions.}
	\label{statistical}
	\vspace{-0.1cm} %
\end{figure}

\subsection{Multi-Subject Image Generation}
We then delve into a more intricate scenario: multi-subject, subject-driven image generation. We scrutinize the quality of multi-subject generation by utilizing all possible combinations (a total of 105 pairs) formed from the 15 subjects described in Section~\secref{Personalized}, allocating 21 prompts to each pair for evaluation. Considering that PortraitBooth was trained on a single-subject dataset, we incorporated the MultiDiffusion~\cite{bar2023multidiffusion} generation method, which combines multiple reference diffusion generation processes with shared parameters, to generate images in different regions during inference. Tab.~\ref{exps_multi} shows a quantitative analysis contrasting PortraitBooth with the baseline methods. The results demonstrate that PortraitBooth significantly improves the identity preservation score. Moreover, our prompt consistency is comparable to tuning-based approaches~\cite{gal2022image, kumari2023multi}, but weaker than FastComposer and Custom Diffusion. We attribute this vulnerability may stem from our method’s inclination to give precedence to subject fidelity. The longer test time, compared to FastComposer, is a result of current multi-subject generation method limitations. We anticipate a significant reduction in our multi-subject generation time as these methods evolve. Fig.~\ref{multi-subject} shows the qualitative results of multi-subject personalization comparisons. Please refer to the \textit{supplementary materials} for more visual examples.


\subsection{Expression Editing}
To demonstrate the effectiveness of our approach in terms of facial expression editing, we conduct a series of comparisons against both test-time fine-tuning methods capable of expression editing and those that are not. 
The entire test set comprises $15$ subjects, as mentioned in Section \secref{Personalized}, with each subject assigned $11$ prompts containing emotion-related words. 
The comprehensive results in Tab.~\ref{expression_compare} clearly show that our method significantly outperforms the others.
Fig.~\ref{emotioncmp} presents the experimental comparison results for expression editing, showcasing the versatility of our method.

\begin{table}[t!]
\centering
\resizebox{\linewidth}{!}{
\begin{tabular}{ccc}
	\toprule
	\bf{Method} &\bf{Type} & \bf{Expression Coefficients} $\uparrow$ \\
	\midrule
	Textual-Inversion \cite{gal2022image} &FineTune& 0.158 \\
	Custom Diffusion \cite{kumari2023multi} &FineTune& 0.182\\
	DreamBooth \cite{ruiz2023dreambooth} &FineTune& 0.153 \\
	FastComposer \cite{xiao2023fastcomposer} &One Shot& 0.133 \\
	\midrule
	\bf{PortraitBooth w/o expression control}&One Shot & 0.177  \\
	\bf{PortraitBooth (Ours)}&One Shot & \bf{0.193} \\
	\bottomrule
	\end{tabular}}
	\vspace{-3mm}
	\caption{\textbf{Comparison of facial expression coefficients} between PortraitBooth and other methods.}
	\label{expression_compare}
\end{table}

\begin{table}[t]
	\centering
	\renewcommand{\arraystretch}{0.8}
	\begin{tabular}{ccc}
		\toprule
		\bf{Combination Type} & \bf{Id Pres.} $\uparrow$ & \bf{CLIP-TI} $\uparrow$ \\
		\midrule
		\{\textit{``person"}\} & 0.623 & 0.229  \\
		\{\textit{``he",``she"}\} &0.606 & 0.208 \\
		\{\textit{``man",``woman"}\} &0.657 & 0.245 \\
		\bottomrule
	\end{tabular}
	\vspace{-3mm}
	\caption{\textbf{The impact of embedding enhancement of subject tokens} from different categories.}
	\label{Combination_Type}
\end{table}

\begin{table}[t!]
	\centering
	\begin{tabular}{cccc}
		\toprule
		\bf{Item} & \bf{Method} & \bf{Id Pres.} $\uparrow$ & \bf{CLIP-TI} $\uparrow$ \\
		\midrule
		& PortraitBooth & 0.657 & 0.245  \\
		(a) & w/o STEA & 0.563 & 0.244\\
		(b) & w/o DIP  & 0.638  & 0.239    \\
		(c) & w/o ECAC & 0.632 & 0.235                  \\
		\bottomrule
	\end{tabular}
	\vspace{-3mm}
	\caption{\textbf{Ablation results of three components.}}
	\label{Ab_results}
\end{table}

\subsection{Ablation Study}
\noindent \textbf{Impact of Identity Token.}
After creating prompts for $70,000$ training images, we analyzed the subject identity token for each image. The results, shown in Fig.~\ref{statistical}, revealed three categories of subject words: \{``man",``woman"\}, \{``person"\}, and \{``he",``she"\}. We tested each category's effectiveness after feature enhancement by converting other identity tokens in each prompt to each experiment token. When converting the "person" token, we manually classified gender correctly for alignment. Our findings, presented in Tab.~\ref{Combination_Type}, showed that the \{``man", ``woman"\} category, being more specific, improved subject fidelity and text-image consistency. The category of \{``he", ``she"\}, \{``person"\} was less descriptive and consistent.

\noindent \textbf{Impact of STEA.}
To investigate the influence of target features obtained from a pre-trained face recognition model, we conducted an ablation. When removing the STEA module, we employed CLIP-image-encoder for training and extracting target features to enhance subject text embeddings. The experimental results, as depicted in Tab.~\ref{Ab_results}(a), clearly indicate that utilizing a face feature extractor trained on a large-scale dataset is significantly more effective compared to training the image encoder.

\noindent \textbf{Impact of DIP.}
Tab.~\ref{Ab_results}(b) presents the ablation studies on our proposed DIP.
As the results show, the DIP module has proven beneficial for identity preservation.

\noindent \textbf{Impact of ECAC.}
To let the model focus on semantically relevant subject regions within the cross-attention module, we incorporate the attention map control. Tab.~\ref{Ab_results}(c) indicates this operation delivers a substantial performance improvement for identity preservation as well as prompt consistency.
Besides, as shown in Tab.~\ref{expression_compare}, even when our cross-attention control mechanism does not constrain the expression terms, we still achieve satisfactory results in facial expression editing. This further demonstrates the effectiveness of our method in decoupling identity and attributes.

\begin{figure}
	\centering
	\includegraphics[width=0.5\textwidth]{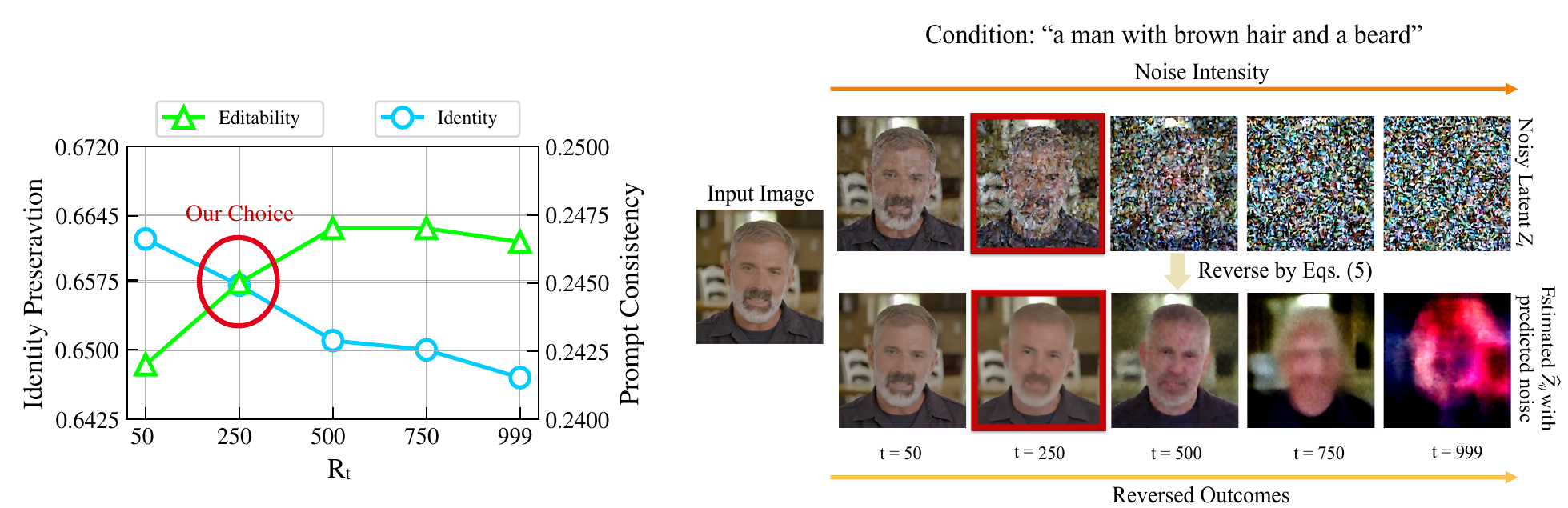}
	\vspace{-8mm}
	\caption{\textbf{Effects of using different upper limit of timesteps} for one-step reverse (left), \textbf{visualization of noise addition} at different timesteps $t$ and denoising (right).}
	\label{different_t_reverse}
	\vspace{-0.5cm}
\end{figure}

\begin{figure}[t]
	\centering
	\includegraphics[width=0.5\textwidth]{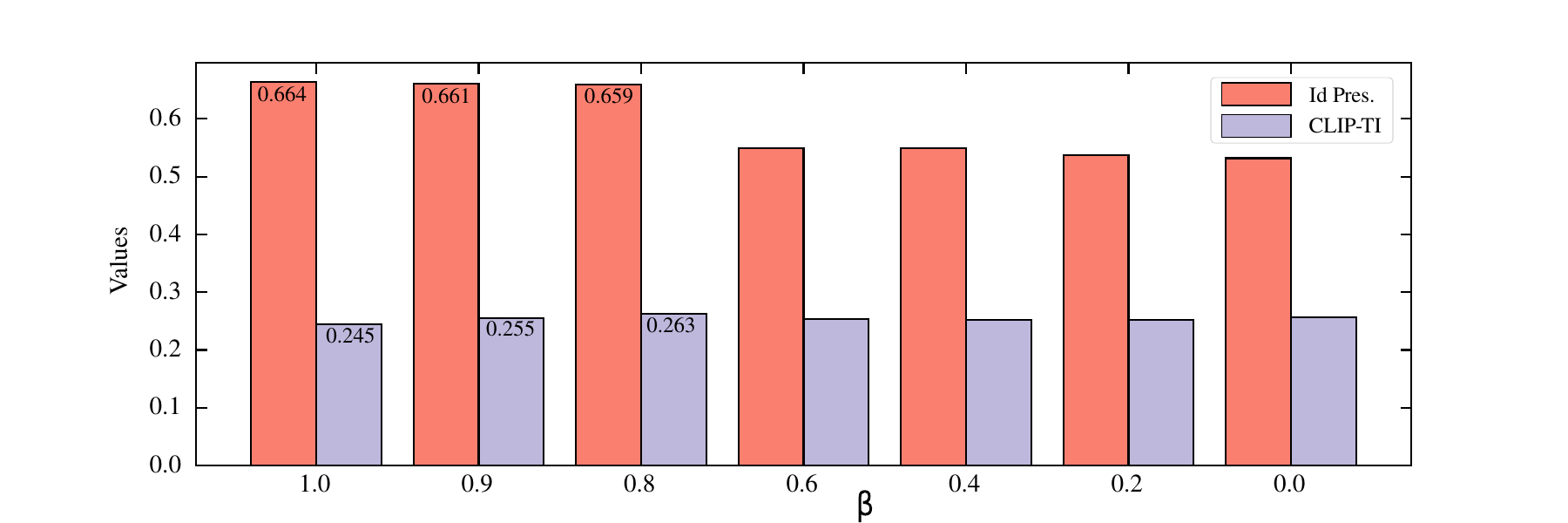}
	\vspace{-8mm}
	\caption{\textbf{The impact of truncating cross-attention} only with different values of $\beta$.}
	\label{beta}
	\vspace{-0.2cm} %
\end{figure}

\noindent \textbf{Hyperparameter $R_t$.}
As shown on the left side in Fig.~\ref{different_t_reverse}, when $R_t$ grows, the model trades off identity preservation for improved editability. We select $250$ as the optimal $R_t$ value, as it provides a good balance. The right side of the figure illustrates the visual results.


\noindent \textbf{Hyperparameter $\beta$ and $\gamma$.}
We studied the balance between identity preservation and editability by solely adjusting the  $\beta$ in the truncation process, keeping $\gamma$ at 0, to minimize their impact. As shown in the Fig.~\ref{beta}, when $\beta$ is in the range of [0.8, 1], the difference in identity preservation is not significant, but there is a noticeable change in editability.
However, when $\beta$ is less than 0.8, there is a sudden jump in identity preservation. We believe this is because the enhanced face embeddings have a significant effect only on the facial region. Tab.~\ref{facepersonmask} confirms our hypothesis. Therefore, we chose $\beta$ as 0.8 as our hyperparameter. Similarly, for the hyperparameter $\gamma$
, we conducted experiments with $\gamma$ values of 0.1 and 0.2. In the Tab.~\ref{betagamma}, we found that while the difference in identity preservation is not significant between the two values, there is a substantial difference in editability. This is because facial responses include not only expressions but also features like facial hair and accessories, \textit{etc}.
Hence, we select $\gamma$ as 0.1 as our hyperparameter.

\begin{table}[t]
	\centering
	\renewcommand{\arraystretch}{0.8}
	\begin{tabular}{ccc}
		\toprule
		\bf{Mask Type} & \bf{Id Pres.} $\uparrow$ & \bf{CLIP-TI} $\uparrow$ \\
		\midrule
		Face Mask & 0.657 & 0.245  \\
		Person Mask &0.623 & 0.229 \\
		\bottomrule
	\end{tabular}
	\vspace{-3mm}
	\caption{\textbf{Impact of different types of masks.} ``Face Mask" refers to the segmentation of only the facial area, while ``Person Mask" refers to the segmentation of the entire person's body.}
	\label{facepersonmask}
	
\end{table}

\begin{table}[t]
	\centering
	\renewcommand{\arraystretch}{0.8}
	\begin{tabular}{ccc}
		\toprule
		\bf{$\gamma$ and $\beta$ combination} & \bf{Id Pres.} $\uparrow$ & \bf{CLIP-TI} $\uparrow$ \\
		\midrule
		$\beta = 0.8$, $\gamma = 0.1$ & 0.657 & 0.245  \\
		$\beta = 0.8$, $\gamma = 0.2$ &0.652 & 0.223 \\
		\bottomrule
	\end{tabular}
	\vspace{-3mm}
	\caption{\textbf{The influence of different combinations of $\beta$ and $\gamma$.}}
	\label{betagamma}
	\vspace{-0.2cm}
\end{table}

\vspace{-1mm}
\section{Conclusion}
In the portrait personalization field, we face the core challenge of proposing an efficient, low training cost, and high identity preserving portrait personalization framework.
In this paper, we propose PortraitBooth, an efficient one-shot text-to-portrait generation framework, that leverages Subject Text Embedding Augmentation and Dynamic Identity Preservation to achieve robust identity preservation, using Emotion-aware Cross-Attention Control to achieve expression editing, respectively. Experimental results demonstrate the superiority of PortraitBooth over the state-of-the-art methods, both quantitatively and qualitatively.
We hope that our PortraitBooth will serve as a baseline work in this field, which can be followed, reproduced, and optimized by all research institutions.

\noindent\textbf{Limitations.} 
Our work is primarily centered around human-centric portrait imagery. However, we are confident that broadening our dataset to encompass images from diverse categories will substantially boost our model's ability to generate full-body representations.

{
    \small
    \bibliographystyle{ieeenat_fullname}
    \bibliography{main}
}

\end{document}